\documentclass[journal]{IEEEtran}

\usepackage[labelformat=simple]{subcaption}

\usepackage[font=small]{caption}
\usepackage{graphicx}
\usepackage{tabularx}
\usepackage{times}
\usepackage{amsmath}
\usepackage{amssymb}
\usepackage{comment}
\usepackage{url}
\usepackage{multirow}
\usepackage{colortbl}
\usepackage{booktabs}
\usepackage[table,xcdraw]{xcolor}
\usepackage[normalem]{ulem}
\newcolumntype{x}[1]{>{\centering\arraybackslash}p{#1pt}}
\newcolumntype{y}[1]{>{\raggedright\arraybackslash}p{#1pt}}
\newcolumntype{z}[1]{>{\raggedleft\arraybackslash}p{#1pt}}
\usepackage{xcolor}
\usepackage{pifont}
\usepackage{array}
\usepackage{xspace}
\usepackage{amsmath}
\usepackage{nicematrix}

\usepackage{makecell}
\usepackage{threeparttable}
\usepackage{booktabs}
\usepackage{booktabs}
\usepackage{makecell}
\usepackage{threeparttable}
\usepackage{graphicx}
\usepackage{amssymb}

\usepackage{adjustbox}
\usepackage{algorithm}
\usepackage{algorithmic}

\makeatletter
\let\NAT@parse\undefined
\makeatother
\usepackage{xcolor}
\definecolor{MKTwoSix}{RGB}{33,102,172}
\usepackage[colorlinks=true, linkcolor=black, citecolor=black, urlcolor=MKTwoSix]{hyperref}
\usepackage{booktabs}
\usepackage{threeparttable}

\useunder{\uline}{\ul}{}
\useunder{\cellcolor[HTML]{EFEFEF}}{\hl}{}

\newcommand{\cmark}{\ding{51}}
\newcommand{\xmark}{\ding{55}}

\definecolor{best}{rgb}{1.0, 0.6, 0}
\definecolor{best2}{rgb}{1.0, 0.8, 0.6}

\title{CrossTracer: Cross-Embodiment Navigation via VLA Model Reasoning and Trace Residuals Adapting}

\author{}
\author{
    Yao Wang$^{1,2}$,
    Siyuan Wang$^{2}$,
    Zhirui Sun$^{3}$,
    Wenzheng Chi$^{4}$,
    Liang Lin$^{1}$,
    Jiankun Wang$^{2,\dagger}$,
    Wenjun Xu$^{1,\dagger}$

    \thanks{$^1$ Peng Cheng Laboratory}
    \thanks{$^2$ Southern University of Science and Technology}
    \thanks{$^3$ Innovation Investment Research Institute}
    \thanks{$^4$ Soochow University}
    \thanks{$^\dagger$ Corresponding author. Email: {\tt\footnotesize wangjk@sustech.edu.cn, xuwj@pcl.ac.cn}}
}

\begin{document}

\maketitle
\pagestyle{plain}

\begin{abstract}
Vision-language-action (VLA) models provide strong semantic priors for robot navigation, but they often ignore embodiment-specific mobility constraints. A path that is semantically plausible for one robot may be physically infeasible for another. We propose CrossTracer, a hierarchical framework for cross-embodiment navigation through adaptive trace residuals. CrossTracer represents navigation plans as normalized image-plane waypoints, forming a unified pixel-space interface between semantic reasoning and physical grounding. First, Vision-Language Trace Proposer (VL-Tracer) adapts a pretrained VLA model to predict an initial navigation trace from egocentric observations and flexible goal specifications. Second, CE-Adapter refines this trace by predicting embodiment-conditioned residual corrections from visual traversability cues, robot identity, and the initial trace. To train the refinement module without costly manual annotation, Cross-Embodiment RRT* (CE-RRT*) converts panoptic segmentation into robot-conditioned traversability cost maps and generates cost-minimizing pixel-space traces. We evaluate CrossTracer on the NaviTrace benchmark, which tests whether a model can generate embodiment-consistent navigation traces from egocentric observations, language instructions, and robot embodiment types. CrossTracer achieves a total score of 45.68, outperforming the strongest evaluated general-purpose baseline, Gemini-2.5-Pro, by 10.01 points, corresponding to a 28.1\% relative improvement. Real-world deployment on wheeled and legged robots further shows improved navigation success and execution efficiency.
\end{abstract}


\section{Introduction}
\begin{figure}[t]
    \centering
    \includegraphics[width=\columnwidth]{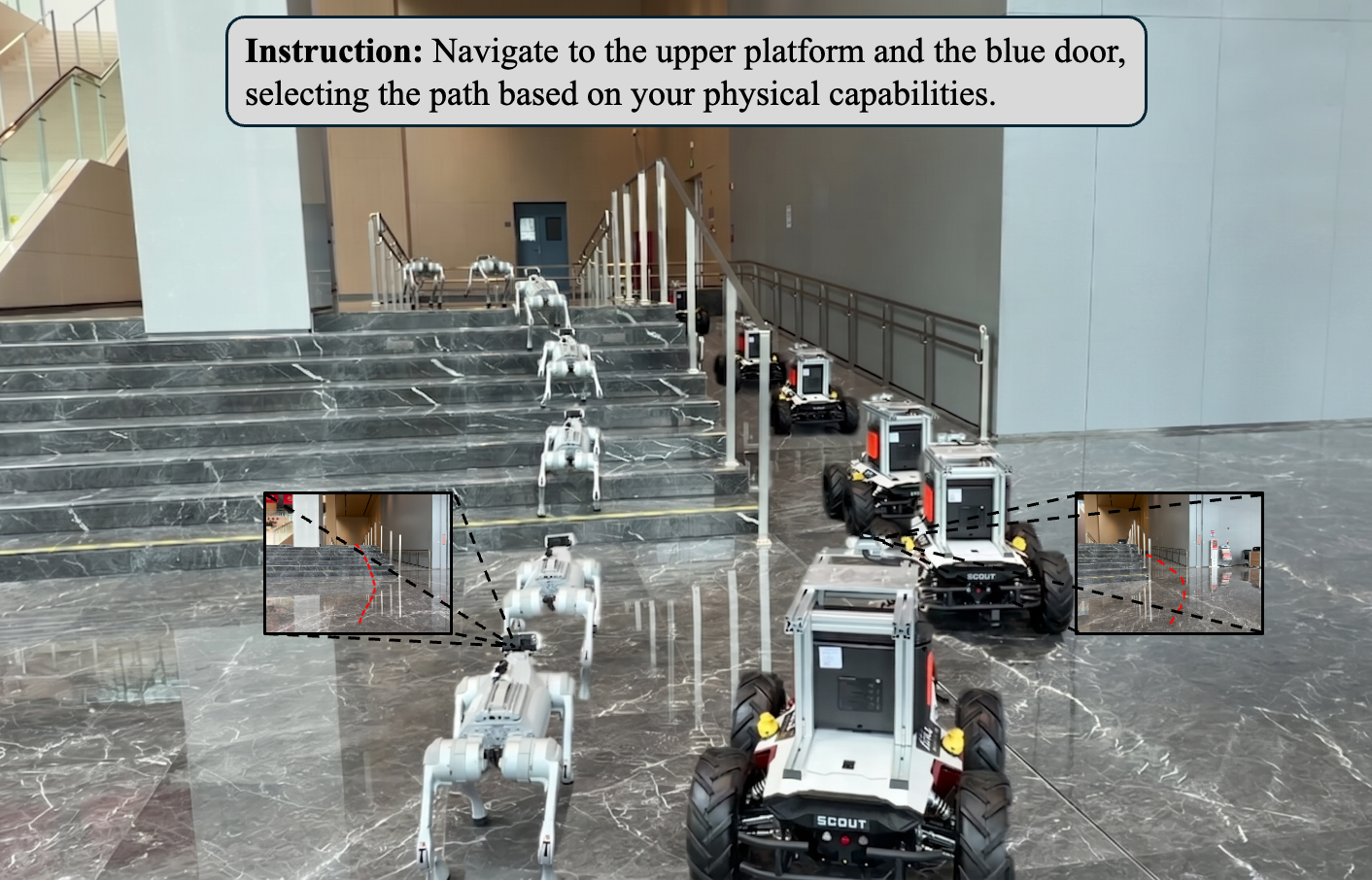}
    \vspace{-0.4cm}
    \caption{
    Embodiment-aware navigation with heterogeneous robots.
    For the same goal, legged and wheeled robots may follow different feasible paths due to distinct mobility constraints.
    The lower panels show their corresponding first-person observations, where red dashed curves indicate the planned pixel-space traces.
    }
    \label{fig:1}
    \vspace{-0.4cm}
\end{figure}

Real-world robot navigation requires a policy to understand high-level semantic goals while respecting the physical constraints of the robot in the execution process. Recent vision-language-action (VLA) models provide strong semantic priors for embodied decision making and navigation~\cite{rt1,rt2,openvla,omnivla}, but their outputs are often insufficiently grounded in robot-specific mobility. The same instruction and observation may imply different feasible routes for different platforms: a legged robot may traverse rough terrain or small height changes, whereas a wheeled robot may need to detour around them, as illustrated in Fig.~\ref{fig:1}. Classical navigation systems and local planners can encode geometric safety and robot constraints~\cite{dwa,teb,rrtstar}, but they typically lack the open-vocabulary goal understanding and semantic flexibility of foundation models. This mismatch brings an embodiment gap between semantic navigation intent and physically executable motion on heterogeneous robots.

Existing approaches address this gap from different directions, but limitations still remain. End-to-end VLA policies~\cite{omnivla,cast,uninavid,navila} map observations and goals directly to actions or trajectories, which can entangle semantic reasoning, embodiment constraints, and control in a single model. Such coupling makes it difficult to adapt the same semantic plan to robots with different traversability profiles. Hierarchical navigation methods partially alleviate this issue by separating high-level reasoning from lower-level grounding. For example, recent pixel-space navigation systems~\cite{vamos} use vision-language models to propose candidate paths and then select or score them according to embodiment-specific affordances. However, candidate selection can only choose among a finite set of proposals and may fail when all candidates contain local infeasible segments. More generally, training embodiment-aware refinement models requires supervision that is costly to obtain across diverse scenes and robot platforms.

Our key insight is that a 2D pixel-space navigation trace can serve as a unified intermediate representation between semantic intent and embodiment-aware physical grounding. A pixel-space trace preserves the scene-level structure inferred by a VLA model from an egocentric observation and a flexible goal specification, while avoiding premature commitment to the low-level control space of any particular robot. At the same time, because the trace is spatially aligned with visual traversability cues in the image, it can be refined by an embodiment-conditioned module that reasons about which region is feasible for a particular robot. This representation therefore enables semantic proposal and physical adaptation to be decoupled while still communicating through a common coordinate interface.

To address the current gap, we propose CrossTracer, a hierarchical VLA framework for cross-embodiment navigation through adaptive trace residuals. CrossTracer first uses Vision-Language Trace Proposer (VL-Tracer), adapted from a pretrained VLA model, to generate an initial embodiment-agnostic semantic trace from an egocentric RGB observation and flexible goal inputs, including language, pixel pose, or both. It then applies CE-Adapter, an embodiment-aware refinement module that predicts residual corrections to the initial trace. The adapter conditions on robot identity and visual traversability cues through robot embeddings, Feature-wise Linear Modulation (FiLM) layers, and trace-to-visual cross-attention, allowing it to preserve the semantic intent of the proposed trace while improving physical feasibility for the target platform.

To train this refinement module without manually annotating embodiment-specific traces, we further introduce Cross-Embodiment RRT* (CE-RRT*), an automated planner-supervised labeling pipeline. CE-RRT* converts panoptic segmentation into robot-conditioned traversability cost maps and applies RRT* planning to synthesize cost-minimizing pixel-space traces. These generated traces provide scalable supervision for learning embodiment-aware residual corrections across different robot types and environments.

Overall, our main contributions are summarized as follows:
{
\renewcommand{\labelenumi}{\arabic{enumi})}
\begin{enumerate}
    \item We propose CrossTracer, a hierarchical cross-embodiment navigation framework that uses normalized pixel-space traces as a unified interface between VLA-based semantic reasoning and robot-conditioned physical grounding.

    \item We develop CE-Adapter, an embodiment-aware refinement module that predicts adaptive trace residuals from visual traversability cues, robot identity, and the proposed trace. To train this module without costly manual annotation, we use CE-RRT*, an automated planner-supervised pipeline that generates embodiment-specific reference traces from semantic segmentation and robot-conditioned cost maps.

    \item We validate CrossTracer on the NaviTrace benchmark and through real-world deployments on wheeled and legged robots, showing that the proposed embodiment-aware trace refinement strategy improves navigation trace quality and physical execution reliability. Experiment videos and more details can be found at \url{https://lilduckkk.github.io/CrossTracer-Nav/}.
\end{enumerate}
}

\section{Related Work}
Research related to CrossTracer can be organized into four main directions: vision-language-action models for navigation, pixel-space trace representations, embodiment-aware navigation, and planner supervision for learning-based navigation. Together, these directions highlight the two central challenges addressed in this work: generating semantically meaningful navigation traces from multimodal goals and adapting them to the physical constraints of heterogeneous robot embodiments.

\subsection{Vision-Language-Action Models for Navigation}

Large-scale robotic foundation models have shown strong potential for language-conditioned robot decision making. RT-1 and RT-2 learn policies that map visual observations and language instructions to robot actions~\cite{rt1,rt2}. RT-X, Open X-Embodiment, Octo, and OpenVLA further scale generalist robot policies using large cross-robot datasets and unified policy architectures~\cite{rtx,octo,openvla}. These models provide useful semantic priors, but most of them are designed for general robot action generation, where semantic reasoning, embodiment constraints, and control are often learned within a coupled policy.

For navigation, learning-based policies such as GNM, ViNT, NoMaD, ViKiNG, MBRA, NavDP, and FlowNav learn scalable visual navigation behaviors from egocentric trajectories, diffusion models, model-based relabeling, or flow matching~\cite{gnm,vint,nomad,viking,mbra,navdp,flownav}. These methods improve goal-directed navigation, but many of them focus on image goals, geometric goals, or action-level navigation policies rather than flexible multimodal goal specification. Recent VLM and VLA navigation systems extend navigation beyond fixed geometric targets by supporting language instructions, image goals, 2D poses, or their combinations. LeLaN learns language-conditioned object navigation from in-the-wild videos~\cite{lelan}, NaVILA combines a VLA module with locomotion skills for legged robot navigation~\cite{navila}, and OmniVLA supports goal images, 2D poses, language prompts, and multimodal combinations~\cite{omnivla}. These systems improve semantic goal understanding and multimodal goal interfaces, whereas CrossTracer uses a VLA-derived module only to propose a semantic pixel trace and leaves embodiment-dependent adaptation to a separate refinement module.

\subsection{Pixel-Space Trace Representations}

Mapless navigation aims to produce feasible navigation directions or trajectories directly from onboard observations, without relying on a prebuilt global map. Classical systems often use geometric perception, local traversability maps, or local planners such as DWA and TEB~\cite{dwa,teb}. Learning-based methods such as MTG generate trajectories with traversability and coverage constraints~\cite{mtg}, while VL-TGS combines trajectory generation with VLM-based selection for mapless outdoor navigation~\cite{vltgs}. CoNVOI uses VLM reasoning over visually marked navigable regions to generate context-aware reference paths~\cite{convoi}. These approaches show that image-aligned spatial representations can support navigation decisions, but they do not directly study residual adaptation of a semantic trace for different robot embodiments.

Pixel-space traces have recently become an effective representation for evaluating and guiding VLM-based navigation. NaviTrace defines a benchmark where a model receives an egocentric image, an instruction, and an embodiment type, and then outputs a 2D navigation trace in image space~\cite{navitrace}. VAMOS uses image-space candidate paths as an interface between a high-level planner and an embodiment-specific affordance model, which evaluates and reranks the candidates~\cite{vamos}. In contrast, CrossTracer does not only select from a fixed candidate set. It treats the pixel trace as a continuous refinement target, proposing an initial semantic trace and then learning residual corrections in the same image plane.

\subsection{Embodiment-Aware Navigation}

Cross-embodiment navigation requires a policy to handle heterogeneous morphologies, mobility limits, and execution constraints. X-Nav trains expert policies over randomly generated embodiments and distills them into a transformer policy for mobile robot navigation~\cite{xnav}. X-Mobility studies end-to-end generalizable navigation through world modeling~\cite{xmobility}. These approaches improve transfer across robot platforms, but they mainly learn navigation behavior in action or control spaces.

Residual learning and generative modeling are also used to improve navigation policies under additional constraints or transfer settings. COMPASS adapts a mobility policy to diverse embodiments through residual reinforcement learning and then distills specialist policies into an embodiment-conditioned policy~\cite{compass}. CE-Nav decouples geometric reasoning and robot dynamic adaptation by training a flow-based velocity expert from planner-generated data, followed by a dynamics-aware refiner~\cite{cenav}. NavDP learns a navigation diffusion policy with privileged simulation guidance for sim-to-real transfer across robot platforms~\cite{navdp}, while FlowNav and FLUX use flow-based generative modeling for efficient navigation policy learning~\cite{flownav,flux}. Unlike these methods, which mainly adapt policies in action, velocity, or control space, CrossTracer performs embodiment-aware adaptation directly in pixel-space by refining a semantic trace produced by an embodiment-agnostic VLA proposer. This design allows semantic goal understanding to remain separate from robot-conditioned traversability refinement.

\subsection{Planner Supervision for Learning-Based Navigation}

Classical planning methods provide interpretable geometric constraints and remain widely used in robot navigation. Dijkstra and A* search for low-cost paths on discrete graphs~\cite{dijkstra,astar}, while RRT and RRT* handle continuous spaces through sampling-based planning~\cite{rrt,rrtstar}. Local planners such as DWA, TEB, and MPC are commonly used for obstacle avoidance and trajectory tracking~\cite{dwa,teb,mpc}. Perception-aware navigation systems further construct traversability or cost maps from LiDAR, elevation maps, RGB-D inputs, semantic segmentation, or learned visual traversability~\cite{mtg,vltgs,viplanner,evora,taln}. EVORA models uncertainty-aware traversability for risk-aware off-road autonomy~\cite{evora}, while Traversability-Aware Legged Navigation learns robot-centric traversability from real-world visual and proprioceptive data~\cite{taln}. NeuPAN further couples point-cloud perception with model-based optimization for direct, map-free navigation in cluttered unknown environments~\cite{neupan}. Although these methods improve physical grounding and safety, they usually assume fixed robot capabilities or operate outside open-vocabulary semantic trace generation.

Recent work uses planners, expert policies, or automatically generated labels as scalable supervision for learning-based navigation. CE-Nav uses planner-generated data to train a flow-based velocity expert~\cite{cenav}, MTG learns traversability-aware trajectory generation~\cite{mtg}, COMPASS uses expert policies and policy distillation for cross-embodiment mobility~\cite{compass}, and VAMOS trains an affordance model to evaluate VLM-proposed image-space paths~\cite{vamos}. CE-RRT* follows this general idea of planner supervision, but targets a different learning interface. It converts semantic segmentation into robot-conditioned traversability cost maps and applies RRT* to generate planner-supervised traces directly in pixel space. These traces are then used to train CE-Adapter, enabling CrossTracer to learn embodiment-aware residual corrections without manually annotated traces.

\section{Problem Formulation}
We formulate cross-embodiment navigation as goal-conditioned trace generation in the image plane. Given an egocentric RGB observation, a flexible goal specification, and a target robot embodiment, the task is to predict a sequence of normalized 2D waypoints that describes a feasible navigation trace for that robot. The trace is not a low-level control command or a full 3D state trajectory. Instead, it is an intermediate spatial representation that connects semantic goal understanding with downstream robot execution.

\begin{figure*}[t]
    \centering
    \includegraphics[width=\linewidth]{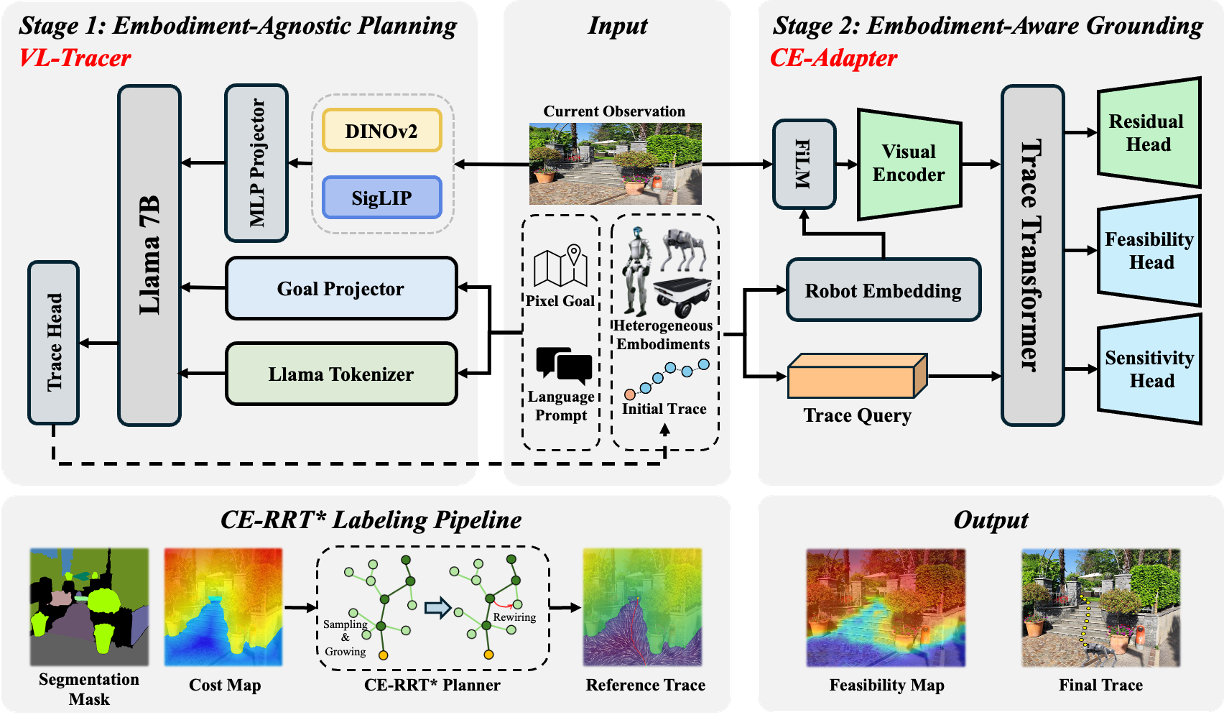}
    \vspace{-0.4cm}
    \caption{
    \textbf{Overview of the proposed CrossTracer framework.}
    CrossTracer follows a two stage design. Stage 1: Embodiment-Agnostic Planning uses the Current Observation, Pixel Goal, and Language Prompt to produce an Initial Trace through Llama 7B and the Trace Head. Stage 2: Embodiment-Aware Grounding incorporates Heterogeneous Embodiments through Robot Embedding and FiLM, then refines the Initial Trace with the Trace Query and Trace Transformer. The Residual Head predicts trace residuals for obtaining the Final Trace, the Feasibility Head predicts the Feasibility Map, and the Sensitivity Head estimates embodiment sensitivity used in the refinement objective. The CE-RRT* Labeling Pipeline generates the Reference Trace from the Segmentation Mask and Cost Map for training supervision.
    }
    \label{fig:framework}
    \vspace{-0.4cm}
\end{figure*}

\subsection{Task Definition}

Let $I \in \mathbb{R}^{H \times W \times 3}$ denote the current egocentric RGB observation, and let $\Omega = \{1,\ldots,W\} \times \{1,\ldots,H\}$ denote the image domain. A goal specification is denoted by $\mathcal{G}$ and may contain a natural language instruction $L$, a target pixel coordinate $P_g=(x_g,y_g)$, or both. The target robot embodiment is denoted by $e \in \mathcal{E}$, where $\mathcal{E}$ is the set of supported embodiments.

The output is an embodiment-conditioned navigation trace
\begin{equation}
    T_e = \{\mathbf{w}_t\}_{t=1}^{N}, \quad
    \mathbf{w}_t = (x_t, y_t) \in [-1,1]^2 .
\end{equation}
Each waypoint $\mathbf{w}_t$ represents an image-plane anchor point along the intended navigation path. The normalized coordinate range $[-1,1]^2$ provides a resolution-independent interface. A waypoint can be converted to image coordinates by
\begin{equation}
    u_t = \frac{x_t + 1}{2} W,\quad
    v_t = \frac{y_t + 1}{2} H .
\end{equation}
Following the trace representation used by NaviTrace~\cite{navitrace}, the model predicts a fixed-length sequence of waypoints. In our implementation, we use $N=8$.

\subsection{Embodiment-Conditioned Traversability}

Different robot embodiments can induce different feasible regions in the same observed scene. For example, a legged robot and a wheeled robot may assign different traversal difficulty to stairs, grass, curbs, narrow passages, or rough terrain. We represent the target embodiment by an embedding $\mathbf{z}_e \in \mathbb{R}^{D_{emb}}$, which is learned from data and used by the refinement model to condition visual features and trace corrections.

For analysis and training supervision, we define an embodiment-conditioned traversability cost map
\begin{equation}
    \mathcal{C}_e : \Omega \rightarrow \mathbb{R}_{\ge 0},
\end{equation}
where lower values indicate regions that are easier for embodiment $e$ to traverse, and higher values indicate unsafe or difficult regions. This cost map encodes the physical feasibility of image regions under the mobility profile of the target robot. It is used to generate planner-supervised reference traces and physical cost signals during training. At inference time, the learned model only requires the RGB observation, goal specification, and embodiment identity.

\subsection{Residual Trace Refinement}

We decompose the problem into semantic trace proposal and embodiment-aware trace refinement. A vision-language trace proposer first predicts an initial semantic trace
\begin{equation}
    T_{init} = f_{\phi}(I,\mathcal{G}),
\end{equation}
which captures the high-level navigation intent but is not explicitly conditioned on the target embodiment. The refinement module then predicts a residual correction conditioned on the observation, initial trace, and embodiment:
\begin{equation}
    \Delta T_e = g_{\theta}(I,T_{init},e).
\end{equation}
The final embodiment-conditioned trace is given by
\begin{equation}
    T_e = T_{init} + \Delta T_e .
\end{equation}

This residual formulation preserves the semantic structure of the initial trace while allowing localized corrections for robot-specific traversability. Excessive deviation from the semantic proposal is discouraged through trace imitation, physical cost, and smoothness losses during training, rather than by an explicit hard bound.

During training, planner-generated reference traces are denoted by $T_e^*$. They represent embodiment-conditioned traces with low traversal cost under $\mathcal{C}_e$. The learning objective is to predict a trace $T_e$ that remains close to $T_e^*$ while preserving the semantic intent encoded by $T_{init}$. The construction of $T_e^*$ and the full training objective are described in the methodology section.

\section{Methodology}

\subsection{System Overview}
\label{sec:system_overview}

CrossTracer is a hierarchical framework that separates goal understanding from embodiment-aware physical grounding. As illustrated in Fig.~\ref{fig:framework}, the framework uses normalized pixel-space traces as the interface between VL-Tracer, the Vision-Language Trace Proposer, and CE-Adapter. VL-Tracer converts the visual observation and goal specification into an initial navigation trace, while CE-Adapter adapts this trace according to visual traversability cues and the target robot embodiment.

Given an egocentric RGB observation $I \in \mathbb{R}^{H \times W \times 3}$, a goal specification $\mathcal{G}$, and an embodiment identity $e \in \mathcal{E}$, CrossTracer maps these inputs to an embodiment-conditioned trace:
\begin{equation}
    T_e = \mathcal{M}(I,\mathcal{G},e).
\end{equation}
The mapping is implemented by two modules. VL-Tracer first predicts an initial trace $T_{init}$ in normalized image coordinates. CE-Adapter then takes $I$, $T_{init}$, and $e$ as input, and refines the trace into the final embodiment-conditioned output $T_e$.

The embodiment identity is introduced only in the refinement stage. This prevents VL-Tracer from entangling goal interpretation with platform-dependent traversability, and allows CE-Adapter to focus on localized corrections for the target robot. During training, VL-Tracer is fine-tuned on navigation trace data through LoRA, while CE-Adapter is trained with planner-supervised reference traces generated by CE-RRT*.

Throughout this section, we denote the initial trace by $T_{init} \in [-1,1]^{N \times 2}$, the final embodiment-conditioned trace by $T_e$, the planner-supervised reference trace by $T_e^*$, the robot embedding by $\mathbf{z}_e \in \mathbb{R}^{D_{emb}}$, and the embodiment-conditioned cost map by $\mathcal{C}_e$.

\subsection{VL-Tracer: Vision-Language Trace Proposer}
\label{sec:vl_tracer}

VL-Tracer maps an egocentric observation and a flexible goal specification to an initial navigation trace in normalized image coordinates. We build VL-Tracer on top of the OmniVLA architecture~\cite{omnivla}, which combines visual encoding with multimodal language reasoning for robot navigation. In CrossTracer, the output interface is changed from low-level navigation actions to a fixed-length sequence of 2D waypoints:
\begin{equation}
    T_{init} = \mathcal{F}_{prop}(I,\mathcal{G})
    = \mathcal{H}_{head}\big(\mathcal{H}_{llm}(\mathcal{H}_{enc}(I,\mathcal{G}))\big),
\end{equation}
where $\mathcal{H}_{enc}$ denotes the visual and goal encoder, $\mathcal{H}_{llm}$ denotes the language model backbone, and $\mathcal{H}_{head}$ denotes the trace prediction head.

VL-Tracer supports goal inputs in three forms: a language instruction $L$, a target pixel coordinate $P_g=(x_g,y_g)$, or both. Let $\mathbf{X}_V \in \mathbb{R}^{M_V \times D}$ denote visual tokens extracted from $I$, $\mathbf{X}_L \in \mathbb{R}^{M_L \times D}$ denote language tokens, and $\mathbf{X}_P = \mathrm{MLP}_{pose}(P_g) \in \mathbb{R}^{1 \times D}$ denote the projected goal-pose token. The tokens are concatenated into a single input sequence:
\begin{equation}
    \mathbf{X}_{input} = [\mathbf{X}_V \,\|\, \mathbf{X}_L \,\|\, \mathbf{X}_P].
\end{equation}
Unavailable modalities are removed by a modality mask $\mathbf{m} \in \{0,1\}^{M}$, where $M=M_V+M_L+1$. During training, language and pose inputs are randomly dropped with probability $p_{drop}=0.3$, which improves robustness to incomplete goal specifications. During inference, the mask is set according to the available inputs.

VL-Tracer predicts a trace $T_{init}=\{\hat{\mathbf{w}}_t\}_{t=1}^{N}$, where each waypoint $\hat{\mathbf{w}}_t=(\hat{x}_t,\hat{y}_t)$ lies in $[-1,1]^2$. Pixel-space traces provide an image-aligned interface between visual goal understanding and downstream embodiment-aware refinement. They are not treated as complete 3D trajectories; instead, they represent spatial anchors in the current egocentric view that can be refined and executed by downstream modules.

The trace head maps the hidden states from the language model to 2D waypoint coordinates:
\begin{equation}
    \hat{\mathbf{w}}_t = \tanh\big(\mathrm{MLP}_{head}(\mathbf{h}_t)\big),
    \quad t=1,\ldots,N,
\end{equation}
where $\mathbf{h}_t \in \mathbb{R}^{D}$ is the hidden state associated with the $t$-th prediction token. The $\tanh$ operation constrains the output to normalized image coordinates. Conversion to image coordinates follows
\begin{equation}
    u_t = \frac{\hat{x}_t+1}{2}W,\quad
    v_t = \frac{\hat{y}_t+1}{2}H.
\end{equation}

VL-Tracer is fine-tuned through Low-Rank Adaptation (LoRA)~\cite{lora} on navigation trace data from VAMOS~\cite{vamos}. The backbone weights are frozen, while the LoRA parameters and trace prediction head are optimized. Given a reference trace $T^{gt}=\{\mathbf{w}_t^{gt}\}_{t=1}^{N}$, the training objective is
\begin{equation}
    \mathcal{L}_{VL}
    = \frac{1}{N}\sum_{t=1}^{N}
    \|\hat{\mathbf{w}}_t-\mathbf{w}_t^{gt}\|_2^2
    + \lambda_{smooth}
    \frac{1}{N-1}\sum_{t=1}^{N-1}
    \|\hat{\mathbf{w}}_{t+1}-\hat{\mathbf{w}}_t\|_2^2 .
\end{equation}
The smoothness term discourages abrupt waypoint changes in the initial trace. We set $\lambda_{smooth}=0.01$ in all experiments.

\subsection{CE-RRT*: Automated Trace Generation}
\label{sec:ce_rrt}

CE-RRT* generates planner-supervised reference traces for training CE-Adapter. The motivation is that manually annotating embodiment-specific traces is expensive, especially when the same scene may require different paths for different robot platforms. CE-RRT* addresses this issue by converting semantic segmentation into an embodiment-conditioned traversability cost map, and then running RRT* in the image plane to produce a low-cost reference trace.

Given an image $I$, CE-RRT* first applies Mask2Former~\cite{mask2former} with a ResNet-50 backbone to obtain a panoptic segmentation mask $M$. Each pixel $\mathbf{p} \in \Omega$ is assigned a semantic label $M(\mathbf{p})$. For each embodiment $e$, we define a semantic cost configuration
\begin{equation}
    \mathcal{K}_e = \{\mathcal{S}^{e}_{free}, \mathcal{S}^{e}_{soft}, \rho_e\},
\end{equation}
where $\mathcal{S}^{e}_{free}$ contains categories that can be traversed with low-cost, $\mathcal{S}^{e}_{soft}$ contains categories that are traversable but difficult, and $\rho_e$ maps soft-traversable categories to nonzero base costs. The base cost map is defined as
\begin{equation}
    \mathcal{C}_{base}^{e}(\mathbf{p}) =
    \begin{cases}
    0, & M(\mathbf{p}) \in \mathcal{S}^{e}_{free},\\
    \rho_e(M(\mathbf{p})), & M(\mathbf{p}) \in \mathcal{S}^{e}_{soft},\\
    \mathcal{C}_{obs}, & \mathrm{otherwise}.
    \end{cases}
\end{equation}
Here $\mathcal{C}_{obs}$ is the obstacle threshold used by the planner.

To account for clearance around obstacles, we apply Euclidean distance transforms to the traversability mask. Let
\begin{equation}
    \mathcal{T}_e =
    \{\mathbf{p} \in \Omega \mid
    M(\mathbf{p}) \in \mathcal{S}^{e}_{free}
    \cup \mathcal{S}^{e}_{soft}\}
\end{equation}
denote the traversable region for embodiment $e$. For $\mathbf{p} \in \mathcal{T}_e$, we penalize proximity to non-traversable pixels:
\begin{equation}
    \mathcal{P}_{ext}^{e}(\mathbf{p})
    = \lambda_{ext}
    \max\left(0, 1-\frac{d_{free}^{e}(\mathbf{p})}{d_{max}}\right),
\end{equation}
where $d_{free}^{e}(\mathbf{p})$ is the distance from $\mathbf{p}$ to the nearest pixel outside $\mathcal{T}_e$. For $\mathbf{p} \notin \mathcal{T}_e$, we assign an obstacle penalty:
\begin{equation}
    \mathcal{P}_{int}^{e}(\mathbf{p})
    = \mathcal{C}_{obs}
    + \beta \min\left(1,\frac{d_{obs}^{e}(\mathbf{p})}{d_{int}}\right),
\end{equation}
where $d_{obs}^{e}(\mathbf{p})$ is the distance from $\mathbf{p}$ to the nearest pixel in $\mathcal{T}_e$. The final cost map is
\begin{equation}
    \mathcal{C}_e(\mathbf{p}) =
    \begin{cases}
    \mathcal{C}_{base}^{e}(\mathbf{p}) + \mathcal{P}_{ext}^{e}(\mathbf{p}),
    & \mathbf{p} \in \mathcal{T}_e,\\
    \mathcal{P}_{int}^{e}(\mathbf{p}),
    & \mathbf{p} \notin \mathcal{T}_e.
    \end{cases}
\end{equation}
The cost values are clipped to a fixed numerical range before planning.

\begin{figure}[t]
    \centering
    \includegraphics[width=\linewidth]{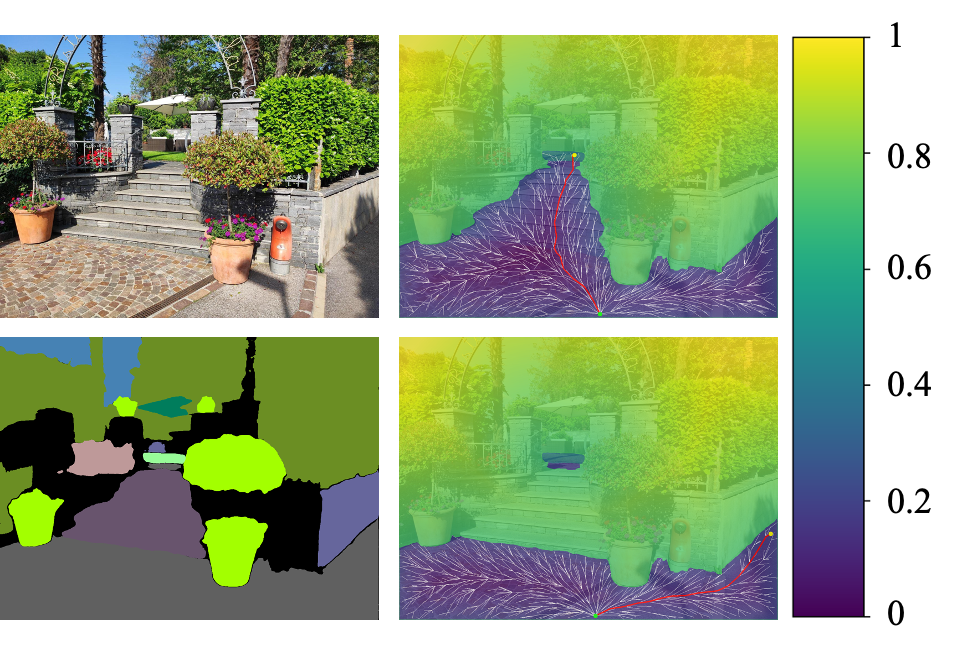}
    \vspace{-0.4cm}
    \caption{
    \textbf{Automated embodiment-specific data generation with CE-RRT*.}
    Given an egocentric RGB observation and a target goal, CE-RRT* first performs semantic segmentation to obtain scene-level traversability cues. The segmentation result is then converted into an embodiment-aware cost map according to the physical constraints of the target robot, such as wheeled or legged locomotion. Based on this cost map, RRT* searches for a collision-free and low-cost trace in pixel space, producing planner-supervised embodiment-specific traces as supervision labels for training CE-Adapter.
    }
    \vspace{-0.4cm}
    \label{fig:3}
\end{figure}

Given the cost map $\mathcal{C}_e$, start point $\mathbf{s}$, and goal point $\mathbf{g}$, CE-RRT* searches for a low-cost path in the image plane. The complete procedure is summarized in Algorithm~\ref{alg:ce_rrt}. For a discrete path $P=(q_1,\ldots,q_m)$, the path cost is
\begin{equation}
    J(P;\mathcal{C}_e)
    =
    \sum_{i=1}^{m-1}
    d(q_i,q_{i+1})\,\omega(q_{i+1}),
\end{equation}
where $d(\cdot,\cdot)$ is Euclidean distance and
\begin{equation}
    \omega(q)=1+\eta\frac{\mathcal{C}_e(q)}{\mathcal{C}_{max}}.
\end{equation}
This cost penalizes both path length and traversal through high-cost regions.

We use RRT*~\cite{rrtstar} to minimize this cost approximately under collision constraints. The planner samples in the $H \times W$ image plane with goal-biased sampling probability $p_{goal}=0.15$, steering step $\Delta_{step}=25$ pixels, search radius $r=60$ pixels, and maximum iteration number $N_{max}=10{,}000$. An edge is accepted only when all sampled points along the segment have cost below $\mathcal{C}_{obs}$. After a path is extracted, it is uniformly resampled into $N$ waypoints and normalized to $[-1,1]^2$, yielding the planner-supervised reference trace $T_e^*$.

The generated $T_e^*$ and cost map $\mathcal{C}_e$ are used only during training. At inference time, CrossTracer does not require semantic segmentation or a precomputed cost map; CE-Adapter learns to infer the needed traversability cues from the RGB observation and embodiment identity.

\begin{algorithm}[t]
\caption{Planner-Supervised Trace Generation via CE-RRT*}
\label{alg:ce_rrt}
\begin{algorithmic}[1]
\REQUIRE Image $I$, goal point $\mathbf{g}$, start point $\mathbf{s}$, embodiment $e$
\ENSURE Reference trace $T_e^*$
\STATE $M \leftarrow \mathrm{Mask2Former}(I)$
\STATE $\mathcal{C}_e \leftarrow \mathrm{GenerateCostMap}(M,e)$
\STATE $\mathcal{V} \leftarrow \{\mathbf{s}\}, \mathcal{A} \leftarrow \emptyset$
\FOR{$i = 1$ \TO $N_{max}$}
    \STATE $q_{rand} \leftarrow \mathrm{Sample}(\Omega,\mathbf{g},p_{goal})$
    \STATE $q_{near} \leftarrow \mathrm{Nearest}(\mathcal{V},q_{rand})$
    \STATE $q_{new} \leftarrow \mathrm{Steer}(q_{near},q_{rand},\Delta_{step})$
    \IF{$\mathrm{CollisionFree}(q_{near},q_{new},\mathcal{C}_e,\mathcal{C}_{obs})$}
        \STATE $\mathcal{Q}_{near} \leftarrow \{q \in \mathcal{V} \mid d(q,q_{new}) \le r\}$
        \STATE $q_{best} \leftarrow \arg\min_{q \in \mathcal{Q}_{near}}
        \mathrm{Cost}(q) + d(q,q_{new})\omega(q_{new})$
        \STATE $\mathcal{V} \leftarrow \mathcal{V} \cup \{q_{new}\}$
        \STATE $\mathcal{A} \leftarrow \mathcal{A} \cup \{(q_{best},q_{new})\}$
        \STATE $\mathrm{Rewire}(\mathcal{Q}_{near},q_{new},\mathcal{C}_e)$
    \ENDIF
\ENDFOR
\STATE $P_e \leftarrow \mathrm{ExtractPath}(\mathcal{V},\mathcal{A},\mathbf{g})$
\STATE $T_e^* \leftarrow \mathrm{ResampleAndNormalize}(P_e,N)$
\RETURN $T_e^*$
\end{algorithmic}
\end{algorithm}

\begin{figure*}[t]
    \centering
    \includegraphics[width=\linewidth]{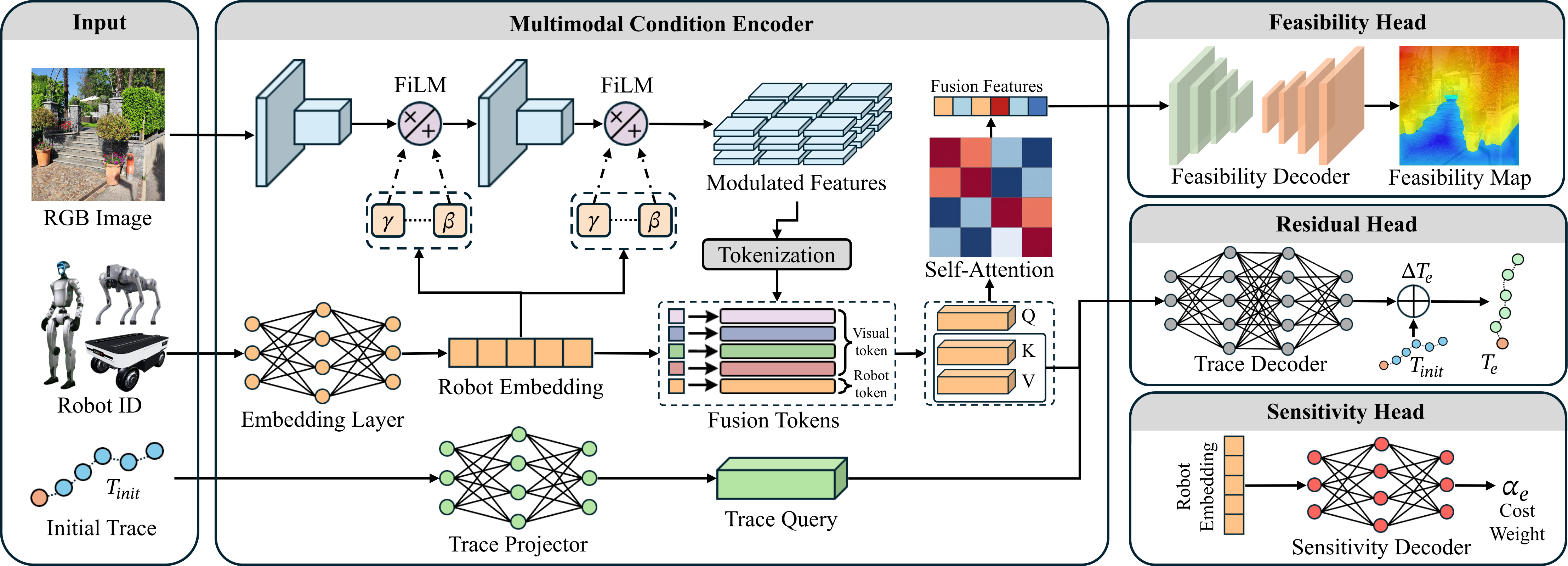}
    \vspace{-0.4cm}
    \caption{
    \textbf{Architecture of CE-Adapter.}
    The Input contains the RGB Image, Robot ID, and Initial Trace. The Multimodal Condition Encoder first maps the Robot ID through the Embedding Layer to obtain the Robot Embedding, which modulates visual features through FiLM. The modulated visual features pass through Tokenization and Self-Attention to form Fusion Tokens and Fusion Features. In parallel, the Initial Trace is encoded by the Trace Projector to form the Trace Query. The Feasibility Head uses the Fusion Features and Feasibility Decoder to predict the Feasibility Map. The Residual Head uses the Trace Decoder to predict $\Delta T_e$ and refine $T_{init}$ into $T_e$. The Sensitivity Head uses the Robot Embedding and Sensitivity Decoder to produce the Cost Weight $\alpha_e$.
    }
    \label{fig:4}
    \vspace{-0.4cm}
\end{figure*}

\begin{figure}[ht]
    \centering
    \includegraphics[width=\columnwidth]{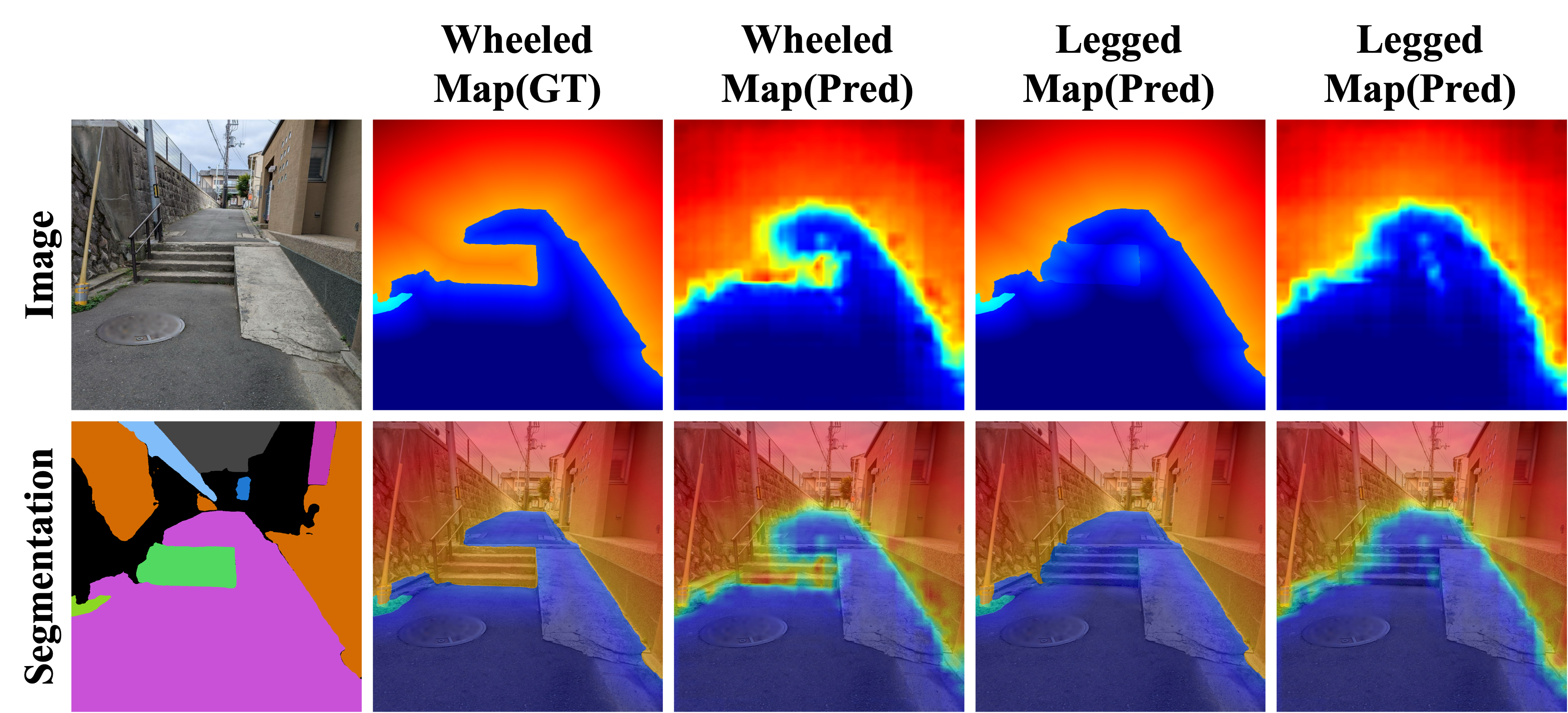}
    \vspace{-0.4cm}
    \caption{
    \textbf{Embodiment-conditioned feasibility prediction.}
    Given the same observation, different robot embodiments induce different traversability distributions. The figure compares RGB observations, the segmentation mask, planner-generated reference maps, and predicted Feasibility Maps for wheeled and legged robots, showing that CE-Adapter learns embodiment-specific terrain affordances from semantic and geometric cues.
    }
    \label{fig:5}
    \vspace{-0.4cm}
\end{figure}

\subsection{CE-Adapter: Adaptive Trace Residual Learning}
\label{sec:ce_adapter}

CE-Adapter refines the initial trace generated by VL-Tracer into an embodiment-conditioned trace. Given the RGB observation $I$, initial trace $T_{init}$, and embodiment identity $e$, the adapter predicts a residual correction $\Delta T_e$ and outputs
\begin{equation}
    T_e = T_{init} + \Delta T_e .
\end{equation}
This residual formulation preserves the global structure of the initial trace while allowing local corrections in regions where the proposed path conflicts with the mobility constraints of the target robot.

The visual branch starts from a pretrained ResNet backbone~\cite{resnet}. To inject robot information into visual perception, the embodiment identity $e$ is mapped to a learnable embedding $\mathbf{z}_e \in \mathbb{R}^{D}$. At each encoder stage $i$, a FiLM layer~\cite{film} predicts affine modulation parameters from $\mathbf{z}_e$ and applies them to the intermediate feature map $F_i \in \mathbb{R}^{C_i \times H_i \times W_i}$:
\begin{equation}
    F_i^{e} = \gamma_i(\mathbf{z}_e) \odot F_i + \beta_i(\mathbf{z}_e),
\end{equation}
where $\gamma_i(\mathbf{z}_e),\beta_i(\mathbf{z}_e) \in \mathbb{R}^{C_i}$ are broadcast over spatial dimensions, and $\odot$ denotes channel-wise multiplication. The FiLM layers are initialized near identity, with $\gamma_i$ initialized around one and $\beta_i$ around zero, so that the adapter starts from the pretrained visual representation and gradually learns embodiment-conditioned feature modulation.

The final modulated feature map is flattened and projected into visual tokens $\mathbf{X}_v \in \mathbb{R}^{L \times D}$. The robot embedding is prepended as a robot token, and positional embeddings are added:
\begin{equation}
    \mathbf{X}_{0}^{e}
    =
    [\mathbf{z}_e \,\|\, \mathbf{X}_v] + \mathbf{E}_{pos}.
\end{equation}
A stack of Transformer blocks processes this sequence and produces embodiment-conditioned visual tokens
\begin{equation}
    \mathbf{X}_{vis}^{e}
    =
    \mathrm{Transformer}(\mathbf{X}_{0}^{e}).
\end{equation}
These tokens encode visual traversability cues under the mobility profile represented by $e$.

The initial trace is then projected into trace queries:
\begin{equation}
    \mathbf{Q}_{trace}
    =
    \mathrm{MLP}_{trace}(T_{init}) + \mathbf{E}_{trace},
\end{equation}
where $\mathbf{Q}_{trace} \in \mathbb{R}^{N \times D}$ and $\mathbf{E}_{trace}$ denotes learnable waypoint position embeddings. The trace queries attend to the embodiment-conditioned visual tokens through multi-head cross-attention:
\begin{equation}
    \mathbf{Z}_{trace}
    =
    \mathrm{CrossAttn}
    \left(
    \mathbf{Q}_{trace}W_Q,
    \mathbf{X}_{vis}^{e}W_K,
    \mathbf{X}_{vis}^{e}W_V
    \right).
\end{equation}

\begin{table*}[t]
\centering
\caption{Performance comparison on the NaviTrace benchmark.}
\label{tab:navitrace}
\begin{adjustbox}{width=\textwidth}
\begin{tabular}{l | c | c | ccccccccccc}
\toprule
Model & \makecell{Open-\\ Source} & $\uparrow$ Total Score & Bicycle & Human & Legged Robot & Wheeled Robot & Accessibility & \makecell{Dynamic \\ Obstacle} & \makecell{Geometric \\ Terrain} & \makecell{Semantic \\ Terrain} & \makecell{Social \\ Norms} & \makecell{Stationary \\ Obstacle} & Visibility \\ \midrule
\rowcolor{gray!15} \multicolumn{14}{l}{\textbf{General Models}} \\
Qwen3-VL-8B-Thinking        & \cmark & -41.30 & -39.92  & -41.64 & -45.31 & -34.58  & -52.25 & -15.77  & -45.16 & -60.62 & -49.55 & -27.22 & -47.33  \\
Claude Sonnet-4.5       & \xmark & 7.36 & 6.12 & 8.31 & 7.11 & 7.14 & 2.15 & 11.16 & 6.22 & -0.01 & -1.91 & 14.83 & 5.86 \\
Qwen3-VL-235B-Thinking & \cmark & 26.24 & 21.86 & 27.31 & 28.38 & 24.12 & 12.21 & 32.20 & 26.82 & 25.17 & 15.45 & 30.82 & 22.63 \\
Gemini-2.5-Pro          & \xmark & 35.67 & 32.15 & 39.32 & 36.46 & 31.07 & 24.23 & 47.90 & 36.80 & 36.01 & 22.91 & 36.79 & 26.39 \\
\midrule
\rowcolor{gray!15} \multicolumn{14}{l}{\textbf{Embodied Models}} \\
MiMo-Embodied-8B-Thinking         & \cmark & -33.55 & -25.32 & -11.48 & -51.80 & -47.34 & -153.30 & -10.17 & -41.26 & -23.95 & -71.89 & -13.55 & -28.00 \\
Robobrain-2.5-8B         & \cmark & 27.96 & 27.67 & 28.23 & 28.99 & 25.87 & 11.41 & 34.30 & 27.87 & 28.31 & 16.60 & 32.52 & 25.74 \\
\midrule

CrossTracer-8B (Ours)             & \cmark & 45.68 & 42.16 & 46.26 & 46.40 & 46.28 & 33.79 & 52.93 & 45.87 & 45.54 & 37.87 & 46.11 & 52.49\\ 
CrossTracer w/o CE-Adapter      & \cmark & 22.56 & 23.94 & 22.64 & 22.47 & 21.41 & -3.18 & 31.68 & 22.38 & 27.53 & 1.28 & 25.23 & 32.27\\

CrossTracer w/ Goal Pose            & \cmark & 63.91 & 60.79 & 64.44 & 64.57 & 64.38 & 53.83 & 71.16 & 63.40 & 61.50 & 63.32 & 63.13 & 64.05\\ \bottomrule
\end{tabular}
\end{adjustbox}
\end{table*}

\begin{figure*}[t]
    \centering
    \includegraphics[width=\linewidth]{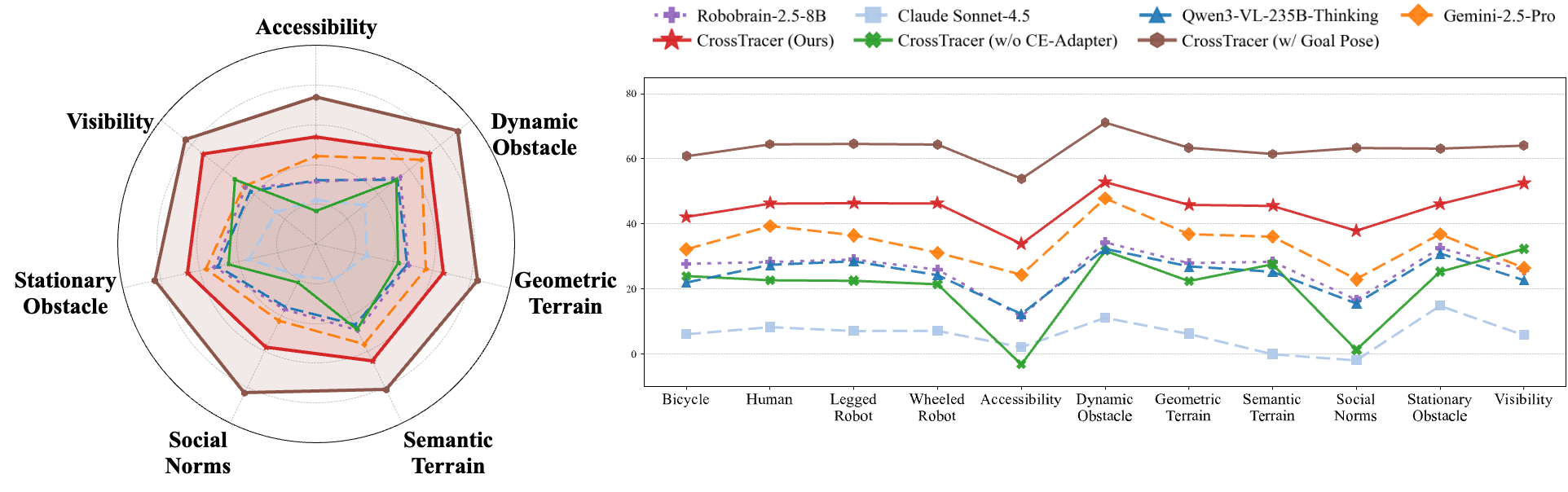}
    \vspace{-0.4cm}
    \caption{
    \textbf{Category-wise performance comparison on the NaviTrace benchmark.}
    The radar chart and line chart compare CrossTracer with representative general-purpose and embodied models across different navigation scenarios, including accessibility, dynamic obstacles, geometric terrain, semantic terrain, social norms, stationary obstacles, and visibility. CrossTracer achieves more balanced and consistently stronger performance, demonstrating the effectiveness of embodiment-aware trace refinement.
    }
    \label{fig:6}
    \vspace{-0.4cm}
\end{figure*}

For compact figure notation, Trace Residual Head, Traversability Reconstruction Head, and Embodiment Sensitivity Head are shown as Residual Head, Feasibility Head, and Sensitivity Head, respectively.
The Trace Residual Head maps the attended trace features to 2D waypoint offsets:
\begin{equation}
    \Delta T_e
    =
    \delta_{max}\tanh\left(\mathrm{MLP}_{res}(\mathbf{Z}_{trace})\right).
\end{equation}
The scaling by $\delta_{max}$ constrains the magnitude of the correction, which helps preserve the goal intent encoded in $T_{init}$ while allowing the adapter to move waypoints away from visually risky or difficult regions.

During training, CE-Adapter also uses two auxiliary heads. The Traversability Reconstruction Head reconstructs the embodiment-conditioned traversability map, represented as the cost map $\mathcal{C}_e$, from the visual tokens and skip features:
\begin{equation}
    \hat{\mathcal{C}}_e =
    \mathrm{Decoder}_{trav}(\mathbf{X}_{vis}^{e}, \{F_i^{e}\}).
\end{equation}
This auxiliary prediction encourages the visual branch to encode terrain affordances that are specific to the target embodiment.

The Embodiment Sensitivity Head predicts a nonnegative coefficient from the robot embedding:
\begin{equation}
    \alpha_e =
    \mathrm{Softplus}(\mathrm{MLP}_{sens}(\mathbf{z}_e)).
\end{equation}
The coefficient modulates the cost-based training term described in Section~\ref{sec:training_objective}, allowing the loss to weight physical cost differently across embodiments.


\subsection{Training Objective and Data Pipeline}
\label{sec:training_objective}

\begin{figure*}[t]
    \centering
    \includegraphics[width=\linewidth]{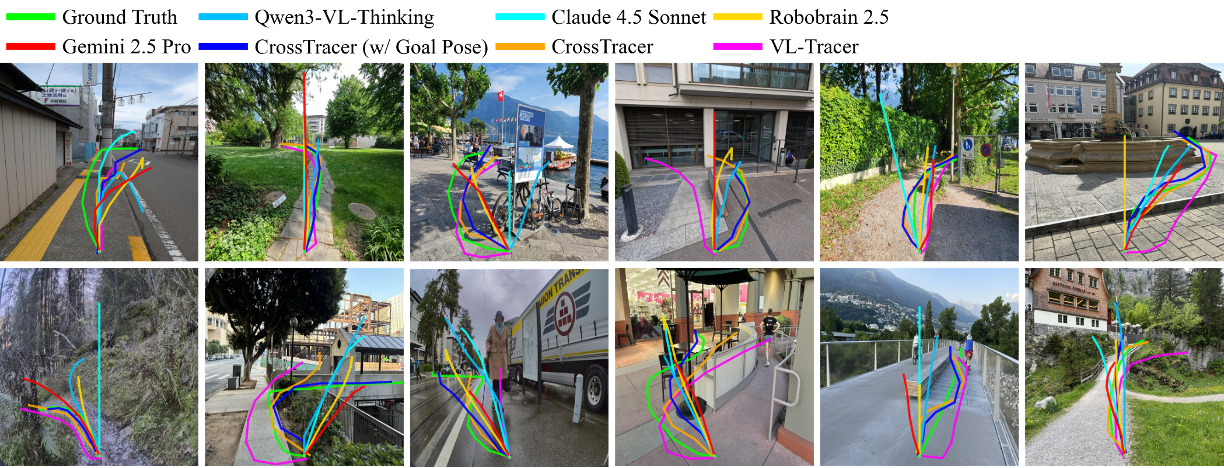}
    \vspace{-0.4cm}
    \caption{
    \textbf{Qualitative results on the NaviTrace benchmark.}
    The figure visualizes pixel-space navigation traces predicted by different models across representative NaviTrace scenarios. These examples cover diverse visual conditions and navigation challenges, including sidewalks, vegetation areas, urban roads, indoor spaces, narrow passages, obstacles, and terrain variations. Compared with baseline models, CrossTracer generates traces that better follow the semantic navigation intent while avoiding physically infeasible regions, demonstrating the benefit of embodiment-aware residual refinement.
    }
    \label{fig:7}
    \vspace{-0.4cm}
\end{figure*}

The training process follows two stages. VL-Tracer is first fine-tuned to produce initial pixel-space traces from visual observations and goal specifications. After this stage, VL-Tracer is frozen. CE-Adapter is then trained to refine the fixed initial traces produced by VL-Tracer with planner-supervised references generated by CE-RRT*. Gradients from the CE-Adapter losses are not back-propagated into VL-Tracer. This separation allows the first stage to learn goal-conditioned trace generation, while the second stage focuses on embodiment-aware residual correction.

CE-Adapter is optimized with a composite objective:
\begin{equation}
    \mathcal{L}_{CE}
    =
    \mathcal{L}_{trace}
    + \lambda_{trav}\mathcal{L}_{trav}
    + \lambda_{cost}\mathcal{L}_{cost}
    + \lambda_{smooth}\mathcal{L}_{smooth}.
\end{equation}
The trace loss supervises the final output of the Trace Residual Head. Given the refined trace $T_e=\{\mathbf{w}_t^e\}_{t=1}^{N}$ and the planner-supervised reference trace $T_e^*=\{\mathbf{w}_t^*\}_{t=1}^{N}$, we define
\begin{equation}
    \mathcal{L}_{trace}
    =
    \frac{1}{N}
    \sum_{t=1}^{N}
    \|\mathbf{w}_t^e-\mathbf{w}_t^*\|_2^2 .
\end{equation}

The Traversability Reconstruction Head is supervised by the embodiment-conditioned cost map $\mathcal{C}_e$ generated by CE-RRT*. Let $\hat{\mathcal{C}}_e$ denote the reconstructed traversability cost map and $\mathcal{P}$ denote the set of image pixels. The reconstruction loss is
\begin{equation}
    \mathcal{L}_{trav}
    =
    \frac{1}{|\mathcal{P}|}
    \sum_{\mathbf{p}\in\mathcal{P}}
    \|\hat{\mathcal{C}}_e(\mathbf{p})-\mathcal{C}_e(\mathbf{p})\|_2^2 .
\end{equation}
This term encourages the visual encoder to learn robot-conditioned terrain affordances instead of relying only on sparse waypoint supervision.

The cost loss penalizes refined waypoints that lie in high-cost regions of the embodiment-conditioned map. The value $\mathcal{C}_e(\mathbf{w}_t^e)$ is obtained by bilinear sampling from the cost map using normalized waypoint coordinates. The Embodiment Sensitivity Head predicts $\alpha_e$, which weights this penalty:
\begin{equation}
    \mathcal{L}_{cost}
    =
    \alpha_e
    \frac{1}{N}
    \sum_{t=1}^{N}
    \mathcal{C}_e(\mathbf{w}_t^e).
\end{equation}
This formulation allows different embodiments to place different emphasis on avoiding high-cost regions during training.

The smoothness loss regularizes the refined trace with second-order finite differences:
\begin{equation}
    \mathcal{L}_{smooth}
    =
    \frac{1}{N-2}
    \sum_{t=1}^{N-2}
    \|(\mathbf{w}_{t+2}^e-\mathbf{w}_{t+1}^e)
    -(\mathbf{w}_{t+1}^e-\mathbf{w}_{t}^e)\|_2^2 .
\end{equation}
This term discourages abrupt direction changes and improves the stability of the predicted trace.

For the data pipeline, VL-Tracer is trained on navigation trace data from VAMOS~\cite{vamos}, because VAMOS provides image-space navigation path annotations that match the pixel-space trace interface used by VL-Tracer. This makes it suitable for learning the embodiment-agnostic semantic trace proposal before CE-Adapter performs robot-conditioned refinement. CE-Adapter is trained on egocentric navigation images annotated by CE-RRT*. For each training image and embodiment, CE-RRT* generates a planner-supervised trace $T_e^*$ and an embodiment-conditioned cost map $\mathcal{C}_e$. During CE-Adapter training, $T_{init}$ is produced by VL-Tracer and then refined by the adapter. During inference, CE-RRT*, semantic segmentation, and cost maps are not required.

We use fixed hyperparameters across all CE-Adapter experiments. Input RGB images and cost maps are resized to $64\times64$, and each trace is represented by $N=8$ normalized waypoints. CE-Adapter is trained with Adam using a learning rate of $1\times10^{-4}$ and a batch size of 64. The loss weights are set to $(\lambda_{trace},\lambda_{trav},\lambda_{cost},\lambda_{smooth})=(1.0,1.0,1.0,0.05)$, where the smoothness term is assigned a smaller weight to regularize abrupt direction changes without suppressing necessary embodiment-conditioned corrections.

\section{Simulation Experiments}

\subsection{Experimental Setup}

\emph{1) Benchmark.}
We evaluate CrossTracer on the NaviTrace benchmark~\cite{navitrace}, which is designed to assess whether vision-language models can generate navigation traces that are consistent with both semantic instructions and embodiment constraints. Each test sample provides an egocentric RGB image, a language instruction, and an embodiment type. The model is required to output a 2D navigation trace in pixel space. We follow the official evaluation protocol and report the total score as the primary metric, where a higher score indicates better performance. We also report scores over the benchmark categories, including bicycle, human, legged robot, wheeled robot, accessibility, dynamic obstacle, geometric terrain, semantic terrain, social norms, stationary obstacle, and visibility.

\emph{2) Baselines.}
We compare CrossTracer with two groups of models. The first group contains general-purpose vision-language models, including Qwen3-VL-8B-Thinking, Claude Sonnet-4.5, Qwen3-VL-235B-Thinking, and Gemini-2.5-Pro. These models provide strong visual reasoning ability but are not specifically trained for embodiment-aware navigation trace generation. The second group contains embodied reasoning models, including MiMo-Embodied-8B-Thinking and Robobrain-2.5-8B. To measure the contribution of the refinement stage, we also report CrossTracer w/o CE-Adapter, which removes the embodiment-aware residual refinement module and uses the trace generated by VL-Tracer directly.

\emph{3) Implementation.}
VL-Tracer is fine-tuned through LoRA on navigation trace data from VAMOS~\cite{vamos}. The backbone\linebreak[3] is frozen, and only the LoRA parameters and trace prediction head are optimized. VL-Tracer fine-tuning is conducted on eight NVIDIA A100 GPUs and takes approximately 48 hours. CE-Adapter is trained on 62k navigation images annotated by CE-RRT*, using the composite objective described in Section~\ref{sec:training_objective}. Training is performed on a single NVIDIA GeForce RTX 4090 GPU and takes approximately 3 hours.

\subsection{NaviTrace Benchmark Results}

\emph{1) Overall performance.}
Table~\ref{tab:navitrace} summarizes the benchmark results. CrossTracer achieves a total score of 45.68, which is the highest score among the evaluated models. Compared with the strongest evaluated general-purpose baseline, Gemini-2.5-Pro, CrossTracer improves the total score from 35.67 to 45.68, corresponding to a relative gain of 28\%. It also outperforms the strongest evaluated embodied model, Robobrain-2.5-8B, by 17.72 points. These results indicate that strong visual reasoning alone is not sufficient for this benchmark, and that explicit embodiment-aware trace refinement provides a clear advantage.

\emph{2) Contribution of CE-Adapter.}
The comparison between CrossTracer and CrossTracer w/o CE-Adapter directly measures the effect of the refinement stage. Removing CE-Adapter reduces the total score from 45.68 to 22.56, a drop of 23.12 points. The improvement is especially large in categories that require physical grounding. Accessibility increases from -3.18 to 33.79, social norms from 1.28 to 37.87, and stationary obstacle from 25.23 to 46.11. These gains suggest that CE-Adapter does more than smooth the initial trace; it learns to correct trace segments that are likely to conflict with embodiment-specific traversability constraints.

\emph{3) Performance across embodiments and scenarios.}
CrossTracer achieves consistent performance across the four embodiment categories, with scores of 42.16 for bicycle, 46.26 for human, 46.40 for legged robot, and 46.28 for wheeled robot. The relatively small variation indicates that the framework does not only fit a single embodiment type. Across scene categories, CrossTracer obtains strong results on dynamic obstacle and visibility, with scores of 52.93 and 52.49, respectively. It also maintains stable performance on geometric terrain and semantic terrain, reaching 45.87 and 45.54. As shown in Fig.~\ref{fig:6}, the performance profile is more balanced than those of the evaluated general-purpose and embodied baselines, especially in categories where physical feasibility and scene layout strongly affect the trace.

\emph{4) Effect of goal-pose input.}
The goal-pose variant further provides a target pixel coordinate in addition to the language instruction and embodiment type. With this additional geometric signal, CrossTracer reaches a total score of 63.91, compared with 45.68 under the default input setting. This variant is not directly comparable to the language-only setting under identical inputs, but it shows that the pixel-space interface can effectively use precise goal location information when available. The improvement is particularly large in semantic terrain and social norms, where the additional target coordinate helps reduce ambiguity in the requested navigation goal.

\subsection{Visual Analysis of Predicted Traces}

Figure~\ref{fig:7} visualizes predicted pixel-space traces on representative NaviTrace scenes. The examples cover sidewalks, vegetation areas, urban roads, indoor spaces, narrow passages, obstacles, and terrain variations. Across these scenes, CrossTracer produces traces that follow the requested navigation direction while better avoiding regions that are visually unsafe or difficult for the target embodiment.

Compared with general-purpose vision-language models, CrossTracer generates traces with clearer spatial structure and fewer abrupt deviations. Compared with CrossTracer w/o CE-Adapter, the refined traces are more consistent with local traversability cues. In scenes containing obstacles, narrow passages, or terrain changes, CE-Adapter adjusts the initial trace away from high-risk regions while preserving the overall goal direction. These examples support the quantitative finding that embodiment-aware residual refinement improves the physical plausibility of predicted navigation traces.

\section{Real-World Experiments}
\subsection{Deployment Protocol}

We evaluate CrossTracer on two heterogeneous mobile robot platforms, including a wheeled robot and a legged robot. The wheeled platform represents ground robots with limited terrain adaptability, while the legged platform can handle more complex surface changes and small height variations. Both platforms are equipped with an egocentric RGB camera, and the same visual observation interface is used during deployment.

\begin{figure}[t]
    \centering
    \includegraphics[width=\columnwidth]{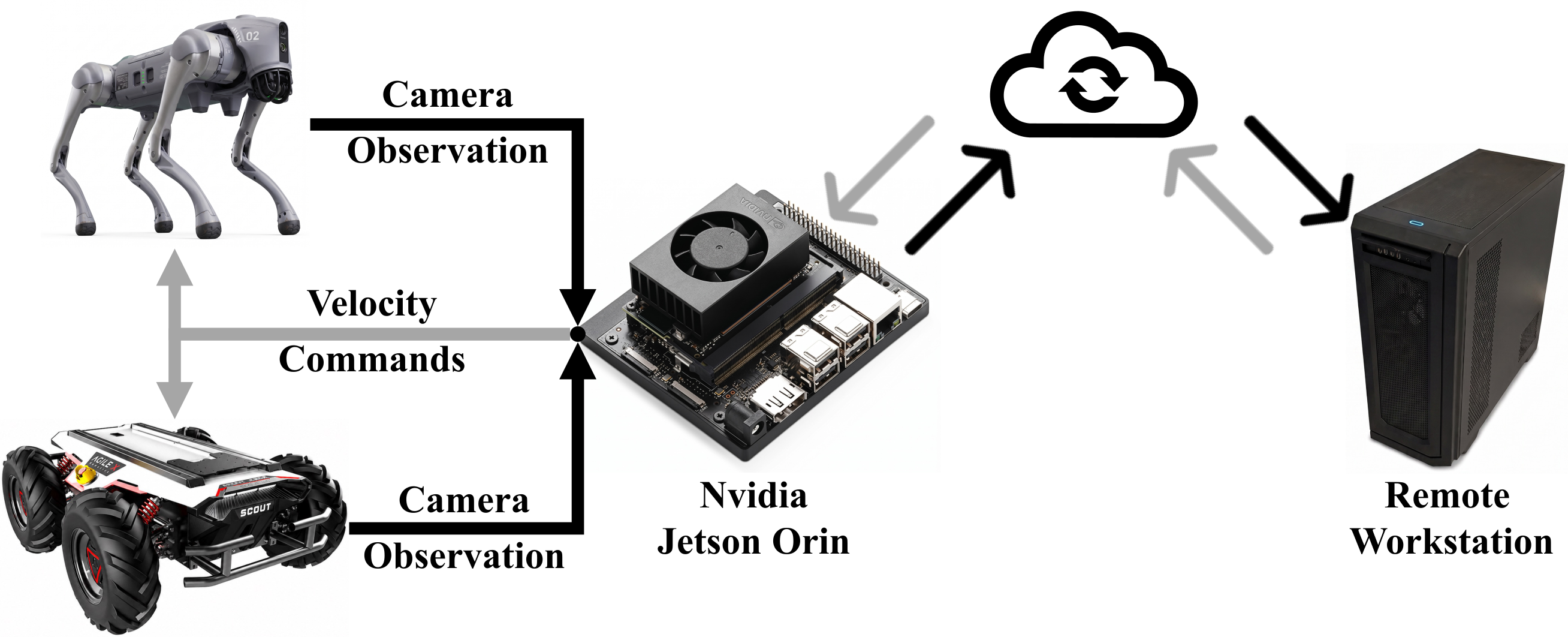}
    \vspace{-0.4cm}
    \caption{
    \textbf{Robot-side and server-side communication pipeline for real-world deployment.}
    The NVIDIA Jetson Orin sends camera observations to a remote workstation for model inference, then converts the returned navigation output into velocity commands for robot execution.
    }
    \label{fig:8}
    \vspace{-0.4cm}
\end{figure}

For each trial, CrossTracer receives the current RGB observation, the navigation instruction, and the robot embodiment identity as input. The model outputs a pixel-space navigation trace in the image plane. This trace is then projected to a sequence of local navigation waypoints and executed by the robot controller in a closed-loop. The low-level controller is kept unchanged across methods, so the comparison focuses on the quality of the predicted navigation trace rather than differences in control implementation.

In our deployment, each robot is equipped with an NVIDIA Jetson Orin, which serves as the onboard computing unit for sensor acquisition, communication, and low-level command execution. The CrossTracer model is deployed on a workstation equipped with an NVIDIA RTX 4090 GPU. During execution, RGB observations captured by the onboard camera are transmitted from the robot to the workstation via WiFi. The workstation performs model inference and sends the predicted pixel-space navigation trace back to the onboard Jetson Orin, where the trace is converted into local navigation waypoints and executed by the robot controller. This setup allows both robot platforms to share the same inference and communication pipeline during real-world deployment.

We compare CrossTracer with OmniVLA under the same robot platforms, navigation instructions, camera observations, and execution protocol. Each method is evaluated from the same start region and toward the same semantic goal. A trial is terminated when the robot reaches the target, collides with an obstacle, deviates from the intended route, or exceeds the maximum execution time. This protocol allows us to assess whether embodiment-aware trace refinement improves physical navigation performance beyond direct vision-language-action prediction. Figure~\ref{fig:8} illustrates the robot-side and server-side communication pipeline used in the real-world experiments.

\begin{figure*}[t]
    \centering
    \includegraphics[width=\linewidth]{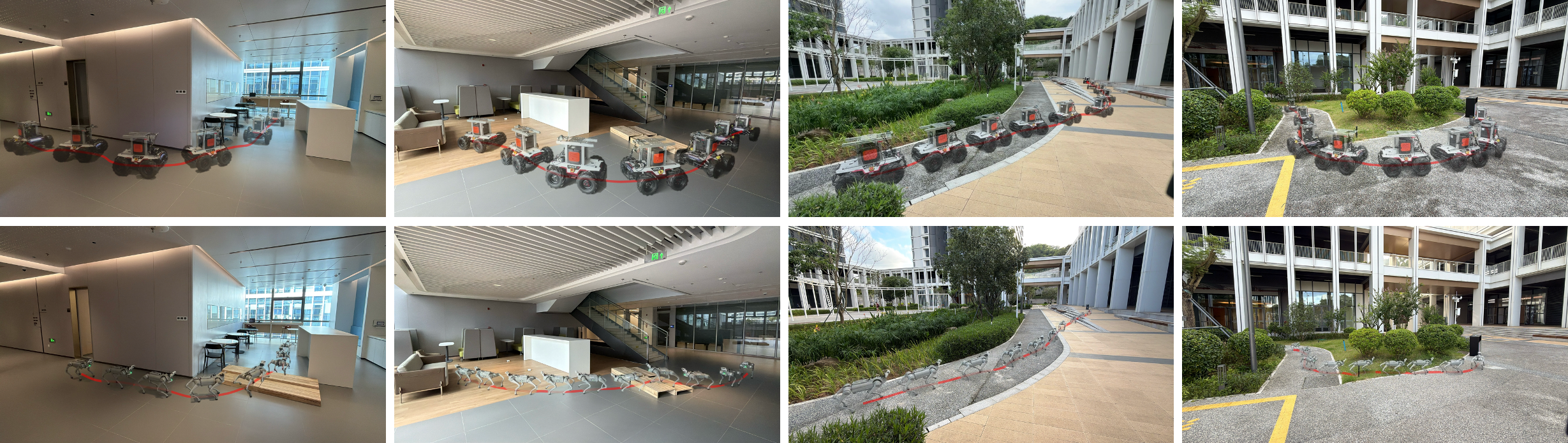}
    \vspace{-0.4cm}
    \caption{
    \textbf{Real-world deployment examples across heterogeneous embodiments.}
    The figure illustrates navigation traces generated for different robot platforms in the same indoor environment. CrossTracer adapts the predicted trace according to the physical characteristics of each embodiment, showing its potential for transferring pixel-space navigation policies across legged and wheeled robots.
    }
    \label{fig:9}
    \vspace{-0.4cm}
\end{figure*}

\begin{figure*}[ht!]
    \centering
    \includegraphics[width=\linewidth]{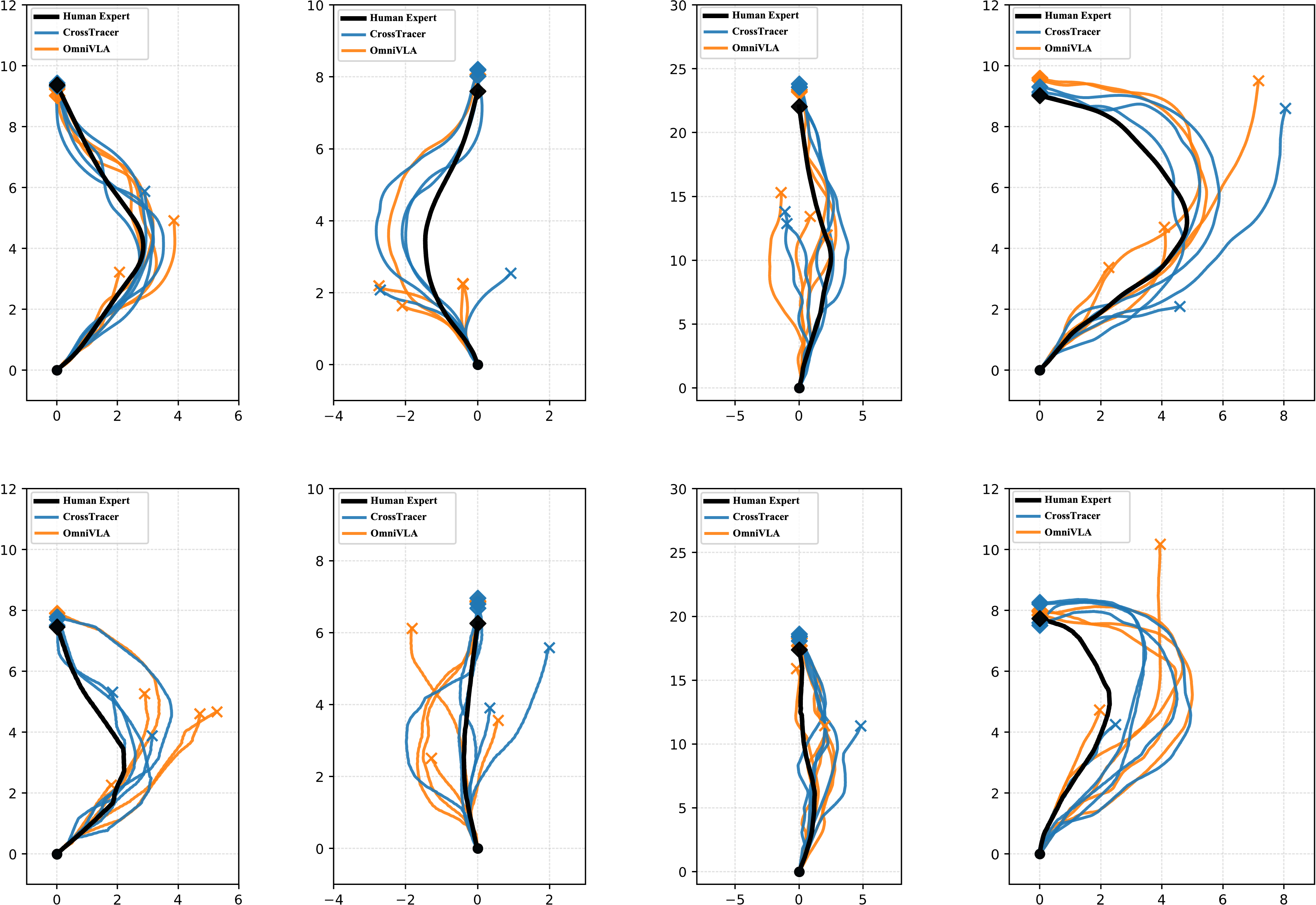}
    \vspace{-0.4cm}
    \caption{
    \textbf{Real-world path comparison in physical deployment.}
    Executed paths from CrossTracer and OmniVLA are compared against human expert reference paths across different real-world navigation trials. Compared with OmniVLA, CrossTracer produces paths that more closely follow the expert references and show smoother convergence near the target, suggesting that residual trace refinement improves physical executability rather than only changing the appearance of the predicted trace.
    }
    \label{fig:10}
    \vspace{-0.4cm}
\end{figure*}

\subsection{Tasks and Evaluation Metrics}

We design four real-world navigation tasks that cover indoor and outdoor deployment scenarios. The indoor tasks require the robot to navigate toward semantic objects in office-like environments, including a white table and a beige sofa located behind a pantry counter. The elevated platform task requires the robot to reach a target sign placed on a second-level platform, which tests whether the generated trace can account for terrain accessibility. The outdoor task requires the robot to follow a long-range instruction with a turning behavior before reaching a black trash bin. These tasks jointly evaluate semantic goal understanding, visual route inference, and embodiment-conditioned feasibility in physical environments.

Each task is repeated five times for each robot platform and each method. A trial is considered successful if the robot reaches within a predefined distance threshold of the target without collision or manual intervention. A trial is marked as failed if the robot collides with obstacles, deviates significantly from the intended route, or exceeds the maximum allowed execution time.

\begin{table*}[t]
\centering
\caption{Real-world navigation performance of OmniVLA and CrossTracer across different robot embodiments. SR, SPL, and STT are reported as ratios, where higher values indicate better performance.}
\label{tab:real_world_results}
\resizebox{\textwidth}{!}{
\begin{tabular}{p{5.4cm} l ccc ccc}
\toprule
\multirow{2}{*}{\textbf{Navigation Instruction}} 
& \multirow{2}{*}{\textbf{Method}}
& \multicolumn{3}{c}{\textbf{Wheeled Robot}}
& \multicolumn{3}{c}{\textbf{Legged Robot}} \\
\cmidrule(lr){3-5} \cmidrule(lr){6-8}
& 
& \textbf{SR$\uparrow$} & \textbf{SPL$\uparrow$} & \textbf{STT$\uparrow$}
& \textbf{SR$\uparrow$} & \textbf{SPL$\uparrow$} & \textbf{STT$\uparrow$} \\
\midrule

\multirow{2}{=}{Navigate to the white table behind the white pantry counter.}
& OmniVLA    & 0.60 & 0.57 & 0.27 & 0.20 & 0.16 & 0.11 \\
& CrossTracer & 0.80 & 0.76 & 0.42 & 0.60 & 0.50 & 0.35 \\

\midrule
\multirow{2}{=}{Navigate to the beige sofa behind the white pantry counter.}
& OmniVLA    & 0.20 & 0.17 & 0.06 & 0.40 & 0.31 & 0.23 \\
& CrossTracer & 0.60 & 0.50 & 0.18 & 0.60 & 0.49 & 0.36 \\

\midrule
\multirow{2}{=}{Navigate to the ``Caution Wet Floor'' sign on the second-level platform.}
& OmniVLA    & 0.40 & 0.37 & 0.23 & 0.60 & 0.35 & 0.40 \\
& CrossTracer & 0.60 & 0.55 & 0.33 & 0.80 & 0.74 & 0.53 \\

\midrule
\multirow{2}{=}{Turn left at the end of the road, then navigate to a black trash bin.}
& OmniVLA    & 0.40 & 0.35 & 0.12 & 0.60 & 0.42 & 0.34 \\
& CrossTracer & 0.60 & 0.54 & 0.27 & 0.80 & 0.58 & 0.46 \\

\midrule
\multirow{2}{=}{Average over all scenarios}
& OmniVLA    & 0.40 & 0.37 & 0.17 & 0.45 & 0.31 & 0.27 \\
& CrossTracer & 0.65 & 0.59 & 0.30 & 0.70 & 0.58 & 0.43 \\

\bottomrule
\end{tabular}
}
\end{table*}

We use three metrics for evaluation: Success Rate (SR), Success-weighted by Path Length (SPL), and Success-weighted by Task Time (STT). SR measures the ratio of successful trials:
\begin{equation}
\mathrm{SR} = \frac{1}{N}\sum_{i=1}^{N} S_i,
\end{equation}
where $N$ is the total number of trials and $S_i \in \{0,1\}$ indicates whether the $i$-th trial succeeds.

SPL evaluates both task success and path efficiency:
\begin{equation}
\mathrm{SPL} =
\frac{1}{N}\sum_{i=1}^{N}
S_i \cdot \frac{L_i^{\mathrm{ref}}}{L_i},
\end{equation}
where $L_i$ is the actual path length executed by the robot, and $L_i^{\mathrm{ref}}$ is the reference path length. Failed trials receive an SPL score of zero because $S_i=0$. In our real-world experiments, $L_i^{\mathrm{ref}}$ is computed from the human expert path.

STT measures the temporal efficiency of successful navigation:
\begin{equation}
\mathrm{STT} =
\frac{1}{N}\sum_{i=1}^{N}
S_i \cdot \frac{T_i^{\mathrm{ref}}}{T_i},
\end{equation}
where $T_i$ is the actual execution time, and $T_i^{\mathrm{ref}}$ is the reference task time. Failed trials receive an STT score of zero because $S_i=0$. Under this definition, a higher STT indicates that the robot reaches the target more reliably and completes the task with higher temporal efficiency.

\subsection{Physical Deployment Results}

Table~\ref{tab:real_world_results} reports the real-world navigation results on the wheeled and legged robots. CrossTracer improves over OmniVLA on both platforms and across all three metrics. On the wheeled robot, CrossTracer increases the average SR from 0.40 to 0.65, SPL from 0.37 to 0.59, and STT from 0.17 to 0.30. On the legged robot, CrossTracer increases the average SR from 0.45 to 0.70, SPL from 0.31 to 0.58, and STT from 0.27 to 0.43. These results indicate that the refined pixel-space traces lead to more reliable and efficient execution in physical environments.

The improvement is consistent across tasks with different navigation requirements. In indoor semantic target reaching, CrossTracer improves the success rate for navigating to both the white table and the beige sofa. In the elevated platform task, the gain is more evident on the legged robot, where the robot can better exploit its terrain adaptability when the predicted trace guides it toward a feasible approach direction. In the outdoor turning task, CrossTracer also improves path efficiency and task-completion efficiency on both platforms. These results suggest that CE-Adapter contributes not only to obstacle avoidance, but also to selecting traces that better match the mobility constraints of the executing robot.

Figure~\ref{fig:9} shows representative deployment examples across heterogeneous robot embodiments. The generated traces preserve the semantic intent of the instruction while adapting to platform-dependent traversability. For the same visual scene and target, CrossTracer can produce different feasible traces for wheeled and legged robots, which provides qualitative evidence that embodiment information is used during trace refinement.

Figure~\ref{fig:10} compares executed paths from CrossTracer and OmniVLA against human expert reference paths. Compared with OmniVLA, CrossTracer produces paths that more closely follow the expert references and show smoother convergence near the target. This qualitative comparison is consistent with the quantitative results in Table~\ref{tab:real_world_results}, suggesting that residual trace refinement improves physical executability rather than only changing the appearance of the predicted trace.

\section{Conclusion}
We presented CrossTracer, a hierarchical navigation framework for embodiment-aware robot navigation with vision-language-action models. The central idea is to decouple semantic route proposal from physical feasibility refinement through a unified pixel-space trace representation. VL-Tracer first predicts an initial navigation trace from egocentric visual observations and flexible goal specifications. CE-Adapter then refines this trace by predicting embodiment-conditioned trace residuals from visual traversability cues, robot identity, and the initial trace. To support scalable training, CE-RRT* automatically generates planner-supervised reference traces by converting panoptic segmentation into robot-conditioned traversability cost maps and applying sampling-based planning.

Experiments on the NaviTrace benchmark show that CrossTracer achieves a total score of 45.68, outperforming the strongest evaluated general-purpose baseline by 28\% and improving over CrossTracer w/o CE-Adapter by 23.12 points. The results indicate that the refinement stage is critical for grounding semantic traces in embodiment-specific traversability constraints. Real-world deployment on wheeled and legged robots further shows that the refined traces improve navigation success, path efficiency, and task-completion efficiency under physical execution.

\section{Limitations and Future Work}
Several limitations remain. The current data generation pipeline relies on panoptic segmentation, so segmentation errors may affect the quality of the traversability cost maps and planner-supervised traces. The cost map configuration for each embodiment is manually specified and may require additional expertise when adapting to new robot platforms. Future work will explore learning embodiment-dependent traversability from robot interaction data, extending the representation to 3D scene structure for handling overhanging obstacles and height discontinuities, and integrating closed-loop replanning for dynamic environments.

\bibliographystyle{IEEEtran} 
\bibliography{IEEEexample}

\end{document}